# Real-Time Helmet Violation Detection in AI City Challenge 2023 with Genetic Algorithm-Enhanced YOLOv5

Elham Soltanikazemi [1,*], Ashwin Dhakal [1], Bijaya Kumar Hatuwal [1], Imad Eddine Toubal [1], Armstrong Aboah [2], Kannappan Palaniappan [1]

[1] Department of Electrical Engineering and Computer Science, University of Missouri, Columbia, MO 65211, USA
[2] Department of Radiology, Northwestern University, Evanston, IL 60208, USA

*Corresponding author: Elham Soltanikazemi (esdft@missouri.edu)

*Abstract*—**This research focuses on real-time surveillance systems as a means for tackling the issue of non-compliance with helmet regulations, a practice that considerably amplifies the risk for motorcycle drivers or riders. Despite the well-established advantages of helmet usage, achieving widespread compliance remains challenging due to diverse contributing factors. To effectively address this concern, real-time monitoring and enforcement of helmet laws have been proposed as a plausible solution. However, previous attempts at real-time helmet violation detection have been hindered by their limited ability to operate in real-time. To overcome this limitation, the current paper introduces a novel real-time helmet violation detection system that utilizes the YOLOv5 single-stage object detection model. This model is trained on the 2023 NVIDIA AI City Challenge 2023 Track 5 dataset, which includes 100 videos used for training and testing, each with an average duration of 20 seconds. The optimal hyperparameters for training the model are determined using genetic algorithms. Additionally, data augmentation and various sampling techniques are implemented to enhance the model's performance. The efficacy of the models is evaluated using precision, recall, and mean Average Precision (mAP) metrics. The results demonstrate impressive precision, recall, and mAP scores of 0.848, 0.599, and 0.641, respectively for the training data. Furthermore, the model achieves notable mAP score of 0.6667 for the test datasets, leading to a commendable 4[th] place rank in the public leaderboard. This innovative approach represents a notable breakthrough in the field and holds immense potential to substantially enhance motorcycle safety. By enabling real-time monitoring and enforcement capabilities, this system has the capacity to contribute towards increased compliance with helmet laws, thereby effectively reducing the risks faced by motorcycle riders and passengers.**

## I. Introduction

Motorcycle accidents pose a significant threat to riders and passengers, with head injuries being the most common and fatal type of injury sustained during such accidents. Wearing a helmet while riding a motorcycle has been shown to reduce the risk of severe head injuries and fatalities by a significant margin. However, despite the known benefits of wearing helmets, compliance with helmet laws remains a major challenge in many parts of the world [1]. Noncompliance with helmet laws may be attributable to several factors, including lack of awareness regarding the importance of helmet use, discomfort caused by helmet wear, and cultural or societal norms that discourage wearing helmets [2]. In some jurisdictions, helmets may be unaffordable or difficult to obtain, exacerbating the issue further. Noncompliance with helmet laws exposes riders and passengers to the danger of serious injuries or fatalities during accidents.

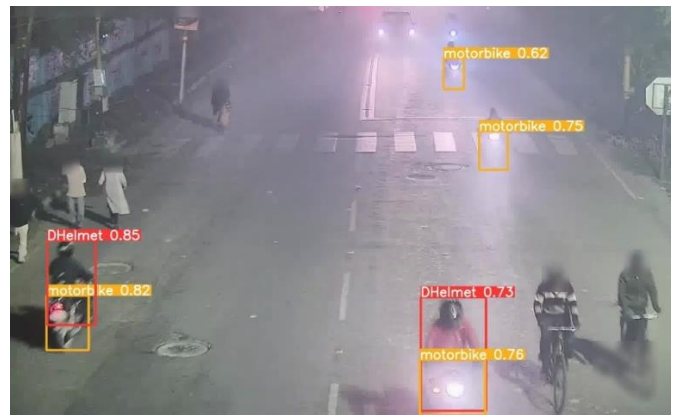

*Figure 1:* An instance of real-time detection of helmet noncompliance in low-light environment.

As a result, there is a growing emphasis on real-time monitoring and enforcement of helmet laws as illustrated in **Figure 1**. Real-time monitoring systems can help identify and discourage non-compliance with helmet laws, ultimately improving motorcycle safety. To address this issue, there have been various attempts to develop systems that can detect helmet use in real-time. Previous approaches have primarily relied on traditional machine learning techniques such as image classification and object detection to identify whether or not a rider is wearing a helmet. While these methods have shown promising results, they are often unable to operate in real-time, which limits their effectiveness in monitoring and enforcing helmet laws. Thus, there is a pressing need for innovative and advanced techniques for real-time helmet violation detection.

In light of these limitations, this paper proposes a real time helmet violation detection system by employing a single-stage object detection model named YOLOv5. In order to produce a highly efficient model, the study employed genetic algorithm to determine the optimal hyperparameters for training. The model was trained on the 2023 NVIDIA AI City Challenge Track 5 dataset. Additionally, certain key data augmentation strategies [3] such as flip, and rotation were implemented to enhance the performance of the model. The efficacy of the developed model was evaluated using the mean average precision (mAP), a widely accepted metric in object detection. It is worth noting that employing the appropriate data augmentation techniques significantly improved the performance of our model.

Overall, our study demonstrates the importance of developing effective real-time helmet violation detection systems that can improve motorcycle safety. Policymakers and law enforcement agencies can leverage our approach to detect and monitor helmet use in real-time, which can ultimately lead to increased compliance with helmet laws and a reduction in motorcycle accidents and fatalities. Our approach represents a significant advancement in the field of real-time helmet violation detection and has the potential to contribute significantly to improving motorcycle safety.

The rest of the paper is organized as follows: Section two provides a review of the relevant literature. In section three, we discuss the data used in this study and the strategies used for preprocessing the data. Section four outlines the helmet detection model and its training process. Section five presents the results obtained from our proposed method and a discussion of those results. Finally, in section six, we provide a conclusion to this study.

## II. RELATED WORK

Previous methods employed for detecting helmet violations can be categorized into two broad groups: statistical or traditional methods and deep learning methods.

### A. Statistical Methods

Classical computer vision techniques for object detection have been a key area of research and development for several decades, predating the emergence of deep learning-based approaches. Early works in image processing explored methods for enhancing, segmenting, and extracting features from images. Dahiya et al. [4] used Gaussian Mixture Models (GMM) for background subtraction since they are generally robust to slight variations in the background, as they can adapt to changes in the background over time. [4] used two hand-engineered features such as scale-invariant feature transform (SIFT) [5], histogram of oriented gradient (HOG) [6], and local binary pattern (LBP) in series, support vector machine (SVM) was used for classification. Dahiya et al. [4]'s technique correctly distinguished between riders and non-riders but struggled to identify the rider's with and without helmets under deplorable conditions. Chiverton et al. [7] introduced an approach that uses SVMs trained on histograms obtained from images of riders' head regions. The riders' heads were isolated and then classified using the trained classifier. Tests show that the classifier performs well when detecting whether riders are wearing helmets or not even with poor illumination. The edge histograms used circular HT causing other circular objects to be misclassified as helmets. Chiu et al. [8] introduce a novel vision-based framework for monitoring riders and their helmets. The proposed system utilizes an advanced occlusion detection and segmentation method based on visual length, visual width, and pixel ratio [9]. This method facilitates the identification and segmentation of various occlusive classes of motorcycles, thereby mitigating the challenges associated with occlusion segmentation and helmet detection complexities.

Kang et al. [10] detect safety helmets efficiently using the ViBe algorithm for background modeling and the C4 framework for locating helmets based on head detection, color feature discrimination, and pedestrian classification. The C4 algorithm works quickly, and color feature discrimination (CFD) determines whether a person is wearing a safety helmet. However, the fixed parameters of C4 and CFD may not perform well in varying environments, requiring parameter adjustments. Waranusast et al. [11] present a system that first detects moving objects using a mixture model algorithm and classifies motorcycles using K-Nearest Neighbor (KNN) classification. Rider heads on the detected motorcycles are counted and segmented by projection profiling. KNN is used to classify the head as wearing a helmet or not based on features from four quadrants of the segmented head region. Silva et al. [12] update the image background using the Adaptive Mixture of Gaussians (AMG). For feature extraction, [12] used hybrid descriptors by combing LBP, HOG, and circular HT descriptors. Classification is achieved using the Naive Bayes classifier (assuming Gaussian distribution of the data and Parzen Window approach), SVMs, and Random Forest (RF) algorithm.

### B. Deep Learning Methods

Deep learning has emerged as a dominant approach for computer vision tasks, supplanting traditional approaches and becoming the preferred method in recent years. Li et al. [13] developed a single-shot multi-box detector (SSD)MobileNet target detection system for real-time safety helmet recognition, however, it exhibited poor generalization for small-pixel helmet images with complicated backgrounds. Chen S. et al. [14] offer an enhanced technique to detect the riders' helmets using the YOLOv5 target detector, attention module, super-resolution reconstruction network, and classifier. High-resolution information for complete interactive fusion optimizes feature extraction in a ladder-type multi-attention network (LMNet)-based model. The new attention module uses channel, location, and spatial information to improve feature representation. Han et al. [15] developed a novel Cross Stage Partial (CSP) module for YOLOv5. The CSP module was designed to reduce information loss and gradient confusion. They used this module in combination with a super-resolution (SR) reconstruction network and the YOLOv5 network to create an end-to-end safety helmet detection model with good accuracy.

Han et al. [16] present a new object detection algorithm based on SSD. It uses a cross-layer attention mechanism for feature refinement and includes a feature pyramid and multiscale perception module for robustness to object scale change. The algorithm also incorporates an adaptive anchor box adjustment method. Li et al. [16] improve the Faster RCNN algorithm for detecting different scales and small objects by using multi-scale training, increasing anchors strategies, and online hard example mining (OHEM) to optimize the model. The improved Faster RCNN detects people and their parts and uses a multi-part combination method with geometric information to detect workers' helmets. The model shows better detection performance for partial occlusion and different-size objects, indicating good generalization and robustness. Zhou et al. [17] suggest a helmet detection algorithm, Attention You Only Look Once (AT-YOLO). They add a channel attention module to the YOLOv3 backbone network for adaptive feature utilization, and a spatial attention module to increase the network's receptive field. [17] use DIoU bounding box regression loss for faster convergence. [18] and [19] use Transformer and attention-based UNET architecture respectively to detect small objects. The optimized training strategy improves network performance without increasing inference costs.

## III. DATA

### A. Data Overview

The dataset is provided by the AI CITY CHALLENGE 2023, where 100 videos are provided in each train and test directory for Track 5 (Detecting Violation of Helmet Rule for Motorcyclists). The videos are in H.264 codec i.e., mp4 format with 10 fps. The videos have an aspect ratio of 16:9 and resolution of $1920 \times 1080$. The average video length of training and testing is 20.01 and 20.0 seconds, respectively.

*Table 1: .The number of instances of classes present in the training data for Track 5 from 2023 AI City Challenge.*

| Class | Number of Instances |
|---|---|
| Motorbike | 29827 |
| DHelmet | 22233 |
| DNoHelmet | 6885 |
| P1Helmet | 97 |
| P1NoHelmet | 4460 |
| P2Helmet | 0 |
| P2NoHelmet | 138 |

Seven different classes are provided as motorbike, DHelmet, DNoHelmet, P1Helmet, P1NoHelmet, P2Helmet, and P2NoHelmet. A file containing the ground truth information with one object instance per line is also provided which has information about video number, frame number, track id, bounding box positions and class value. Among the desired classes, motorbike has the highest number of instances followed by DHelment, DNoHelment, P1NoHelmet, P1Helmet, P2NoHelmet and P2Helmet, respectively, as depicted in **Table 1**. The data is imbalanced because some classes like P2Helmet has no instance, also P1Helmet, P2NoHelmet and P1NoHelmet have very small instances compared to motorbike, DHelment and DNoHelment that can lead to poor predictive performance particularly for minority classes if not handled properly.

### B. Data Preprocessing

To establish the best possible training basis for our object detection model, we conduct several preprocessing steps. As shown in **Figure 2**, we first remove some of highly similar frames belonging to over-represented (majority) class in the training data, as these might significantly bias the training [20] of our object detection model. Then, 95% of background images, which definitely has no object of interest, are discarded since we found that keeping only a tiny fraction of the total background images can help the model to combat the problem of false positives. After these two phases of preprocessing, the number of frames reduces by around 27% to 14556. To improve the quality of remaining frames, we perform data cleansing operations where we remove the false positive, add unannotated objects in frames (false negatives) and resize existing bounding boxes to fit the corresponding objects within frames. Lastly, in order to enhance the generalization ability of our object detection model, we generate some relevant and diverse augmented data using two augmentation techniques including horizontal flip, and rotation.

Furthermore, to build a validation dataset that provides a reliable and unbiased assessment of the model performance when evaluated on testing data, we design a careful train, validation split. More specifically, to ensure that our training and validation data are different, and train, validation subsets sufficiently cover our video dataset, we select all the remaining frames from videos number 91 to 99 as the validation data, and the remaining frames from videos with ID 1 to 90, and also 100 as training data.

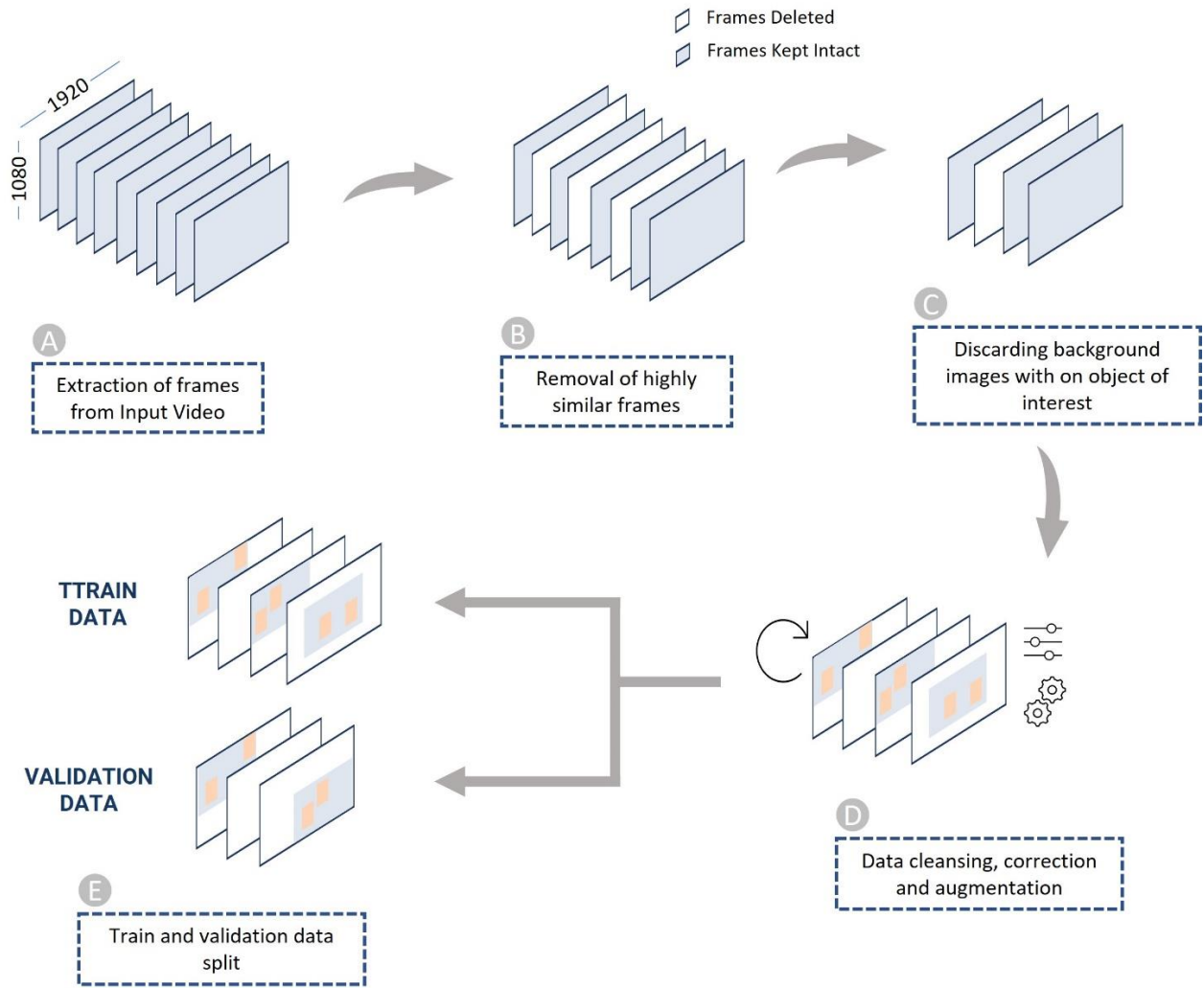

*Figure 2:* Diagrammatic representation of the overall data processing pipeline. (A) Extraction of frames from input videos in H.264 codec format with aspect ratio of 16:9 and resolution 1920 × 1080. (B) Manual removal of highly similar frames. (C) Discarded background images with no object of interest to combat the problem of false positives. (D) Data Cleansing operations by removing false positives, adding unannotated objects in frames, resizing existing bounding boxes, and augmenting data through horizontal flip and rotation. (E) Train and validation split of data for training machine learning model.

## IV. HELMET VIOLATION DETECTION MODEL

This section focuses on constructing a detection model using YOLOv5 that is fine-tuned with the help of the genetic algorithm. Also, a trained YOLOv5 is utilized to design an annotation system to correct labeling errors in the training data.

### A. YOLOv5

YOLOv5 is a single stage object detection algorithm from the YOLO series. The YOLOv5 architecture is composed of three main parts including Backbone, Neck, and Head as shown in **Figure 3 B**. The Backbone contains a convolutional neural network responsible for aggregating and forming image features at contrasting granularities. The architecture's neck fuses the extracted features and generates feature maps to move forward with the prediction task. The head receives the feature maps from the neck and performs the box and class prediction [21]. The backbone network part of YOLOv5, CSPDarknet53, has 29 convolutional layers with kernel size of 33, resulting in a receptive field of size 725 × 725, and a total of 27.6 M parameters. Furthermore, YOLOv5 has a SPP block appended to CSPDarknet53 to widen the receptive field, while keeping the operation speed unaltered. Similarly, PANet is used as a feature fusion to combine low level features with high level features to boost feature richness. The inputs (**Figure 3 A**) were the frames of the video, and the output (**Figure 3 C**) were the predicted labels for helmet usage prediction. YOLOv5 achieved exceptionally good performance through using features such as weighted residual connections, a novel backbone to improve CNN (cross stage partial connections), and a new technique to augment the training data called self-adversarial training.

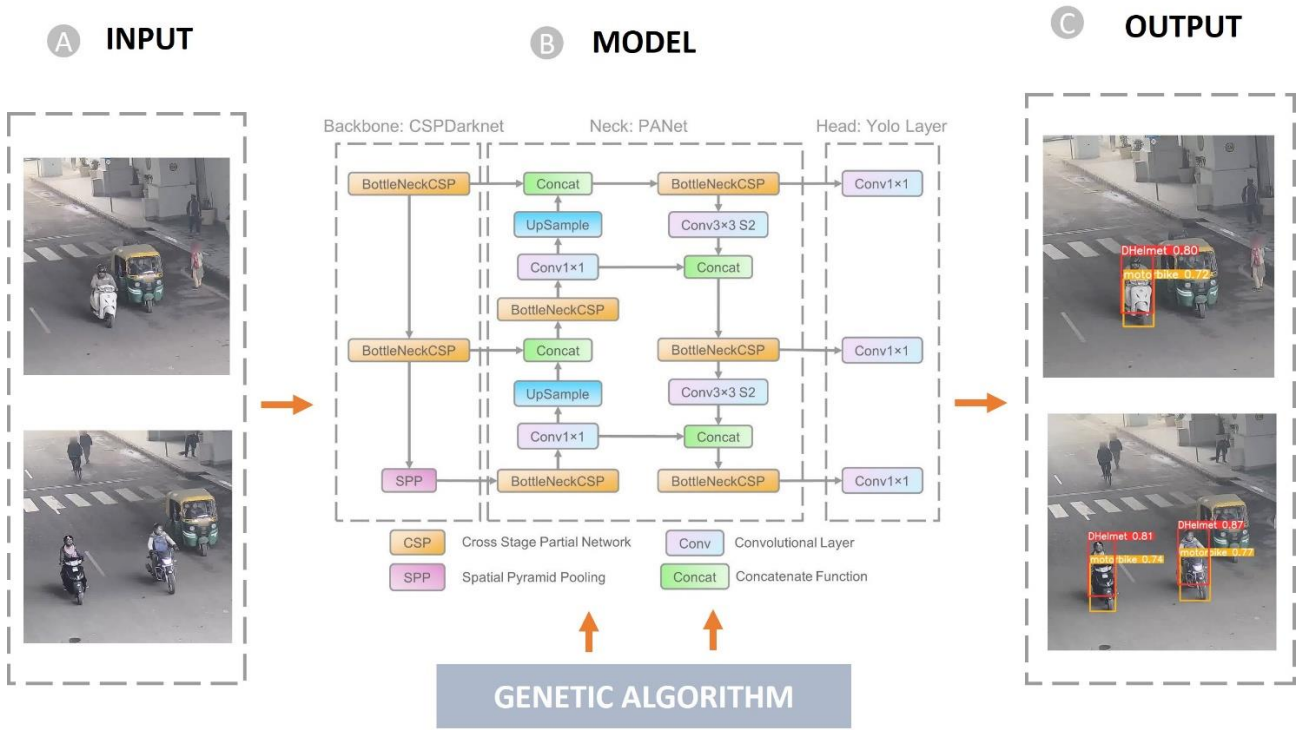

*Figure 3: Overall pipeline of the model architecture. (A) Preprocessed Frames fed to the Backbone of the YOLOV5. (B) YOLOV5 model consisting of Backbone, Neck, and Head with the hyperparameter optimized through genetic algorithm (C) Output of the frame with predicted classes.*

### B. Genetic Algorithm for Hyperparameter Optimization of YOLOv5

Hyperparameter tuning is crucial for the performance of the training algorithm since hyperparameters directly affect the behavior of the algorithm during the training process. However, determining the best combination of values for hyperparameters can be complicated and costly, mainly because of huge search space, and unknown dependencies between parameters. Using traditional approaches like grid search can be exponentially expensive in the number of parameters. Here, we consider the problem of choosing a set of hyperparameter values, which results in the best performance, as an optimization problem, and use the genetic algorithm [22] to solve it. More specifically, we use the genetic algorithm to find the optimal combination for the 30 hyperparameter values of YOLOv5.

*Table 2: Genetic algorithm recommended optimized hyperparameters for YOLOv5 model.*

| Hyperparameters | Representation | Estimated values |
|---|---|---|
| Initial learning rate | lr0 | 0.00936 |
| Final onecyclelr learning rate | lrf | 0.01029 |
| SGD momentum/Adam beta1 | momentum | 0.97177 |
| Optimizer weight decay | weight_decay | 0.00045 |
| Warmup epochs | warmup_epochs | 2.5912 |
| Warmup initial momentum | warmup_momentum | 0.59437 |
| Warmup initial bias lr | warmup_bias_lr | 0.08929 |
| Box loss gain | box | 0.04797 |
| Cls loss gain | cls | 0.6129 |
| Cls bceloss positive_weight | cls_pw | 1.2374 |
| Obj loss gain (scale with pixels) | obj | 0.83451 |
| Obj bceloss positive_weight | obj_pw | 0.64115 |
| Iou training threshold | iou_t | 0.89 |
| Anchor-multiple threshold | anchor_t | 3.9098 |
| Focal loss gamma | fl_gamma | 0 |
| Image HSV-Hue augmentation | hsv_h | 0.01287 |
| Image HSV-Saturation augmentation | hsv_s | 0.80378 |
| Image HSV-Value augmentation | hsv_v | 0.37437 |
| Image rotation (+/- deg) | degrees | 0 |
| Image translation | translate | 0.05457 |
| Image scale | scale | 0.31285 |
| Image shear | shear | 0 |
| Image perspective | perspective | 0 |
| Image flip up-down | flipud | 0 |
| Image flip left-right | fliplr | 0.5 |
| Image mosaic | mosaic | 1 |
| Image mixup | mixup | 0 |
| Segment copy-paste | copy_paste | 0 |
| Anchors per output layer | anchors | 2.1095 |

**Figure 4** shows various plots, each representing different hyperparameter values. The horizontal x-axis displays the range of hyperparameter values, while the vertical y-axis represents its corresponding fitness value. The presence of yellow shading within each plot indicates the maximum value attained for the respective hyperparameter. For instance, the top left corner plot (initial learning rate), demonstrates that a value of 0.00589 yields the highest fitness when utilizing the fitness function of a genetic algorithm.

The genetic algorithm aims at minimizing or maximizing fitness function. We define our fitness function to be the weighted sum of mAP@0.5, and mAP@0.5:0.95, which account for 0.1 and 0.9 of the weight, respectively. Hence, discovering the maximum value of the fitness function serves at finding the best set of hyperparameters. We run the genetic algorithm for 200 generations. Although both crossover and mutation operators can be utilized to create new offspring from selected parents, our genetic algorithm utilizes only the mutation operator, with a probability of 0.8 and variance of 0.04, to produce new children from the best parents. **Table 2** shows some of the hyperparameters found by the genetic algorithm.

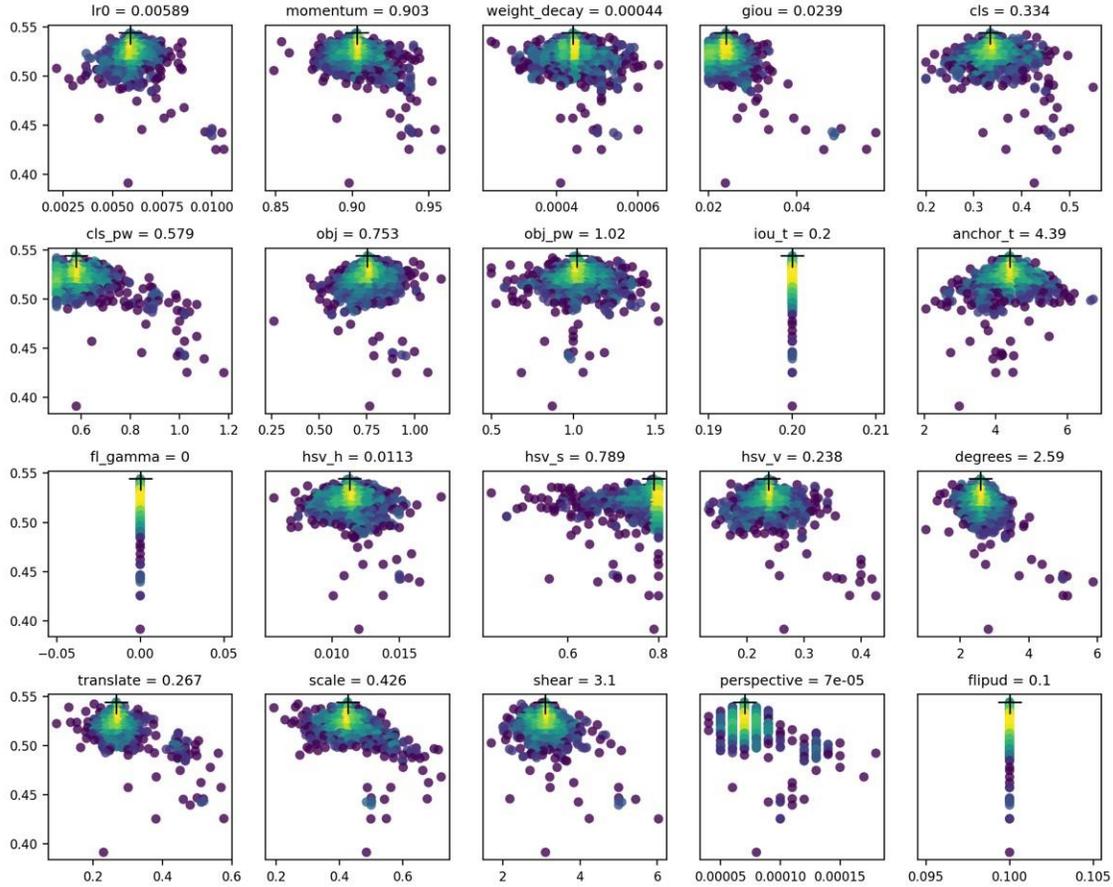

*Figure 4: Depiction of hyperparameter values and their relationship with fitness function. The fitness function values are represented on the y-axis, while the hyperparameter values are displayed on the x-axis. The color yellow highlights the hyperparameter value associated with the highest fitness function value.*

### C. Training The Model

As mentioned earlier, during the training process, we used the frames from videos number 1 to 90, and 100 as training data, and frames from videos number 91 to 99 as validation data. To tackle the class imbalance issue, in the preprocessing phase, we performed undersampling to remove the frames with high similarity from the classes with high records. Similarly, we applied image augmentation techniques [3] such as horizontal flip, and rotation by ±15 degrees to increase the amount of data in minority classes which is known as over sampling. The model was trained for 300 epochs with a SGD optimizer, and a batch size of 32 using the hyperparameters found by the genetic algorithm.

### D. YOLOv5 for Model-assisted Labeling

As discussed above, one of the important steps in the preprocessing phase is to correct labeling mistakes. In order to make this process faster, we use a YOLOv5 model, initially trained on 3609 frames where their labels are first manually corrected, to automatically label the rest of the frames. Some of the automatically generated labels by the model are then manually verified and corrected. Having the corrected labels, we again train a new model used for another round of automatic labeling followed by manual correction. We repeat this process to get correct labels for most of the frames in our training data that are then used for training our final helmet detection model.

## V. RESULTS AND DISCUSSION

The model performance is evaluated according to its ability to correctly detect a total of seven types of classes used in this study including Motorbike, DHelmet, DNoHelmet, P1Helmet, P1NoHelmet, P2Helmet, and P2NoHelmet, corresponding to numbers 1 to 7, respectively.

The evaluation system accepts a txt file, which must have the following format: Video ID, Frame No, Bounding box information, Class ID, and Confidence. Each video is identified with a unique ID from 1 to 100, named Video ID. Frame No indicates the position of a frame in a video. Bounding box information represents the upper left coordinate, width, and height of the box. In addition, Class ID determines the type of the predicted object. Our model is evaluated on the test dataset of Track 5 from 2023 AI City Challenge using Mean Average Precision (mAP).

To optimize our predictions, we employ diverse training methodologies to enhance the performance of our YOLOV5 model in predicting helmet usage among motorbike riders. In our initial approach, we utilized random sampling to select 4,834 frames from a total of 20,000 frames. Each training image was 1920x1080 pixels. The training-validation split was 90:10. Hyperparameters were optimized using genetic algorithm, and precision, recall, and mean average precision (mAP) were recorded on both validation and testing datasets. Test-time augmentation was applied to assess model robustness. This 2nd approach involved uniform sampling of 2 frames per second, resulting in 512x512 pixel images. Like the 1st approach, hyperparameter tuning was conducted to maximize model performance. Continuing with uniform sampling at 2 frames per second, image size was increased to 640x640 pixels to evaluate resolution impact on model performance in the 3rd approach.

Uniform sampling was maintained at 2 frames per second, with images set to the original resolution of 1920x1080 pixels in 4th approach.

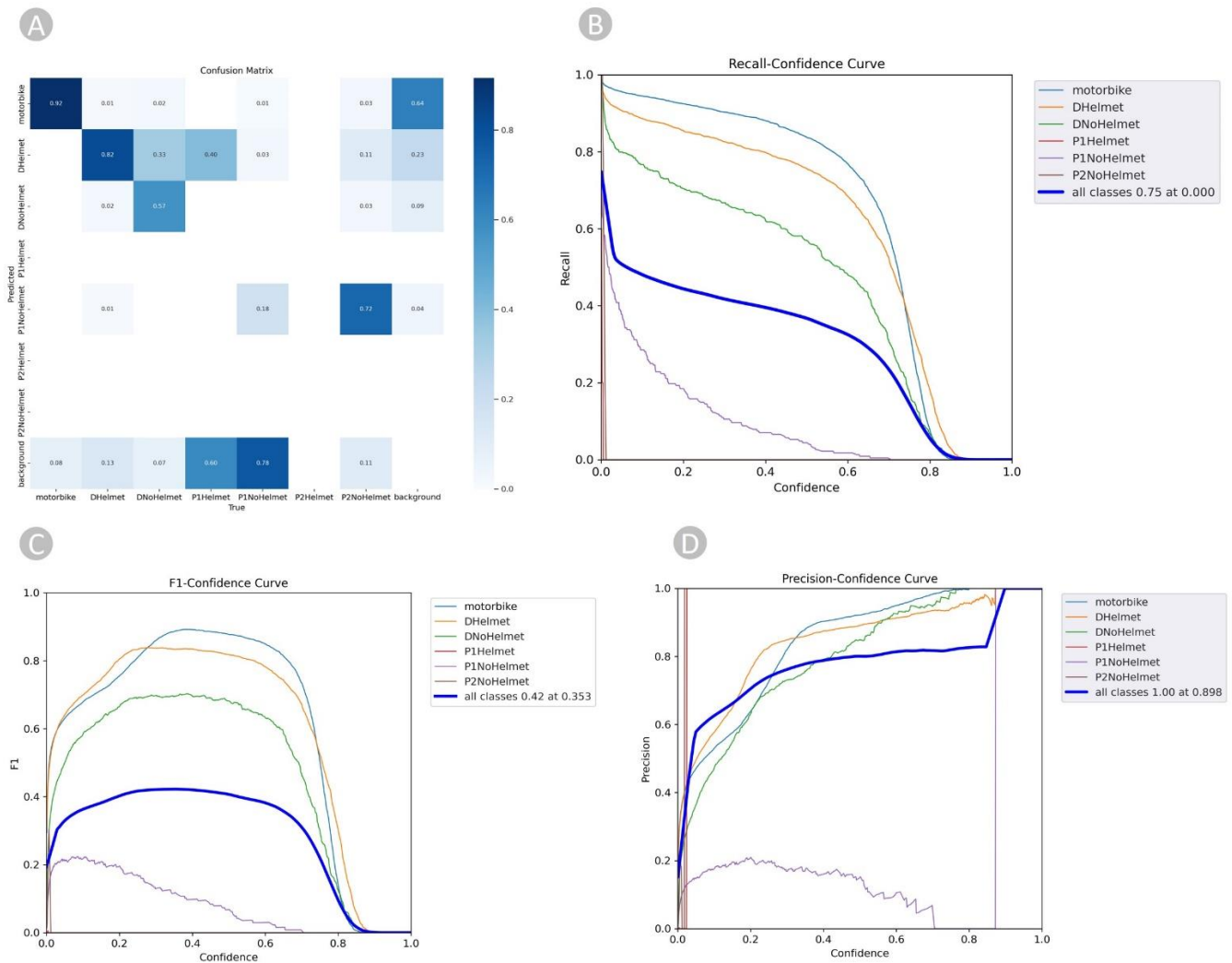

*Figure 5: Prediction metrics and plots. (A) Confusion Matrix representing prediction outcomes. (B) Recall vs confidence plot (C) F1 score vs confidence plot (D) Precision vs confidence plot.*

To address data imbalance, we employed an oversampling technique in the 5th approach. Augmentation was applied to minority classes within 1920x1080 pixel images, maintaining a frame rate of 2 frames per second. Next, to mitigate class imbalance, we undertook under-sampling from the majority class while adhering to 2 frames per second and 1920x1080 pixel images. In the 7th approach, Cross-validation was executed with 10 groups of 100 videos, utilizing 90 videos for training and 10 for validation in each iteration. The ensemble [23] of 10 models was evaluated to determine the most effective approach.

**Figure 5 A** displays a Confusion Matrix representing prediction outcomes. The diagonal elements represent the matches between predictions and actual ground truth values. For instance, in the case of the "motorbike" class, the mAP stands at 0.92, signifying a 92% accuracy in predictions for this class. However, it's important to note that certain instances have null mAP values, as these instances may be absent from the training or validation dataset, thus preventing the model from making predictions for them.

Furthermore, we conducted an assessment of precision, recall, and F1 score across various confidence scores. In **Figure 5 B C D**, each class is represented by a distinct-colored curve. Notably, the best results were achieved with a confidence score of 0.8. To clarify, a confidence score of 0.8 implies that we only consider predictions with an Intersection over Union (IOU) exceeding 0.8, a practice that aids in diminishing the number of false positives in our predictions. In **Figure 5 C**, it's evident that the F1 score reaches its peak at a confidence score of 0.4 but subsequently declines. This pat99tern is due to the fact that when we elevate the confidence threshold, the recall, as depicted in **Figure 5 B**, decreases. In contrast, as we increase the confidence threshold, the precision, as illustrated in **Figure 5 D**, increases.

In addition to the validation metrics and plots, **Figure 6** demonstrates the bounding box, class id, and confidence predicted by our model for some test cases involving different weather conditions and different times of day.

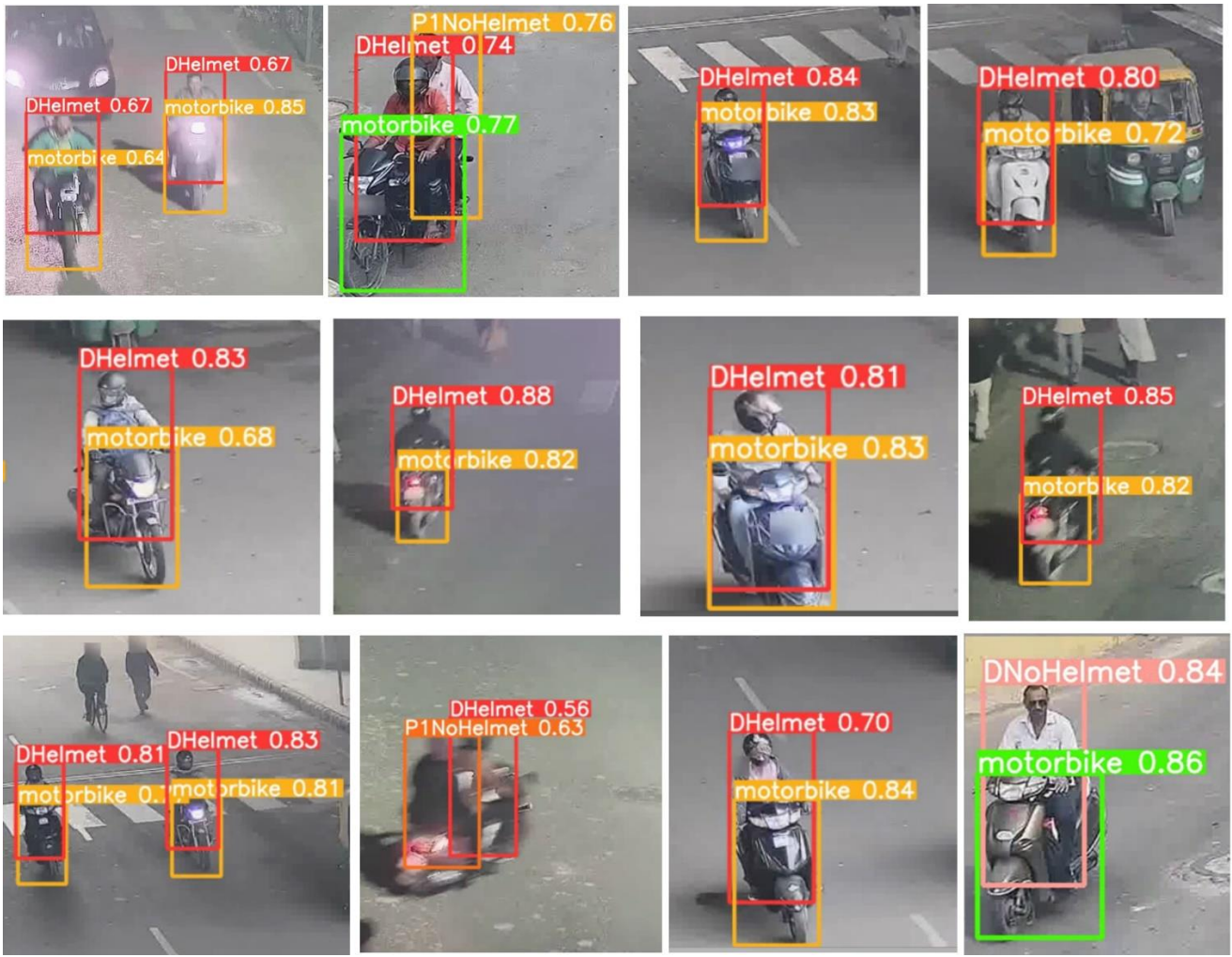

*Figure 6: The prediction results in real time for diverse weather conditions and time. The different bounding box colors correspond to distinct predictive classes.*

**Table 3** below presents the outcomes of our extensive experimentation, highlighting the performance metrics of each training approach. Notably, approach 7, incorporating cross-validation and ensemble learning, emerged as a promising technique, ranking the best in terms of performance metrics.

*Table 3: Comparison of performance metrics for different input data handling approaches.*

| Approaches | Technique to Training Methodologies | Test Time Augmentation | Validation Results | | | Test Results (server) |
|---|---|---|---|---|---|---|
| | | | Precision | Recall | mAP | mAP |
| Approach I | Random sampling | No TTA | 0.775 | 0.398 | 0.471 | 0.3882 |
| | | TTA | 0.753 | 0.428 | 0.479 | 0.3968 |
| Approach II | Uniform sampling (frame rate reduction) | No TTA | 0.702 | 0.424 | 0.562 | 0.4701 |
| | | TTA | 0.678 | 0.454 | 0.56 | 0.4827 |
| Approach III | Uniform sampling with varied resolution | No TTA | 0.706 | 0.454 | 0.614 | 0.4889 |
| | | TTA | 0.683 | 0.454 | 0.627 | 0.4924 |
| Approach IV | Oversampling for imbalanced data | No TTA | 0.762 | 0.569 | 0.6 | 0.5876 |
| | | TTA | 0.66 | 0.46 | 0.604 | 0.6074 |
| Approach V | Oversampling for imbalanced data | No TTA | 0.865 | 0.623 | 0.65 | 0.5864 |
| | | TTA | - | - | - | 0.6213 |
| Approach VI | Under sampling for balanced training | No TTA | 0.853 | 0.577 | 0.602 | 0.5149 |
| | | TTA | - | - | - | 0.6045 |
| Approach VII | Cross-validation for ensemble learning | No TTA | 0.848 | 0.599 | 0.641 | 0.6528 |
| | | TTA | - | - | - | 0.6667 |

Our helmet detection model ranked 4[th] and a score of 0.6667 (see **Table 4**), proving that our model is capable of perfectly detecting the desired objects from many of the videos.

*Table 4: Top 10 teams ranked according to mAP on the AI City Challenge 2023 test dataset.*

| Rank | Team ID | Team Name | Score |
|---|---|---|---|
| 1 | 58 | CTC-AI | 0.8340 |
| 2 | 33 | SKKU Automation Lab | 0.7754 |
| 3 | 37 | SMARTVISION | 0.6997 |
| **4** | **11** | **AIMIZ (ours)** | **0.6667** |
| 5 | 18 | UT He | 0.6422 |
| 6 | 16 | UT NYCU SUNY-Albany | 0.6389 |
| 7 | 45 | UT Chang | 0.6112 |
| 8 | 192 | Legends | 0.5861 |
| 9 | 55 | NYCU- Road Beast | 0.5569 |
| 10 | 145 | WITAI-513 | 0.5474 |

## VI. CONCLUSION

In this study, we developed a helmet violation detection system utilizing a YOLOv5 model fine-tuned by the genetic algorithm. We employed different methods of selecting frames to address the issue of imbalanced data during training. Our model is sufficiently robust to handle real-world low-quality images with complicated backgrounds. Through evaluating our model on a challenging dataset, which provides an accurate representation of real-world images, we proved that our work can positively affect the safety of motorbike drivers and their passengers by automatically detecting the violation of not wearing a helmet. Moving forward, our future work will be on amassing additional training data to enhance the overall performance of our system, ensuring its suitability for real-world implementation.